%% file: BITE.tex
\documentclass[conference, table, xcdraw]{IEEEtran}
\IEEEoverridecommandlockouts
\usepackage[switch]{lineno}
\usepackage{booktabs}
\usepackage{cite}
\usepackage{amsmath,amssymb,amsfonts}
\usepackage{algorithmic}
\usepackage{graphicx}
\usepackage{textcomp}
\usepackage{xcolor}
\usepackage{booktabs}
\usepackage{multirow}
\usepackage{array}
\usepackage{subfig}
\usepackage{siunitx}
\usepackage[colorlinks,
            linkcolor=blue,
            anchorcolor=blue,
            citecolor=blue]{hyperref}
\def\BibTeX{{\rm B\kern-.05em{\sc i\kern-.025em b}\kern-.08em
    T\kern-.1667em\lower.7ex\hbox{E}\kern-.125emX}}
\begin{document}

\title{Bidirectional Time-Frequency Pyramid Network for Enhanced Robust EEG Classification}

\author{
\IEEEauthorblockN{
Jiahui Hong\textsuperscript{1,*},
Siqing Li\textsuperscript{2,*},
Muqing Jian\textsuperscript{3},
and Luming Yang\textsuperscript{2,†}
}
\IEEEauthorblockA{
\textsuperscript{1}\textit{Lanzhou University, Lanzhou, China}\\
\textsuperscript{2}\textit{The Ohio State University, OH, US}\\
\textsuperscript{3}\textit{The Chinese University of Hong Kong, Shenzhen, China}\\
\{hongjiahui0110, siqingli2004\}@gmail.com, muqingjian@link.cuhk.edu.cn, \textsuperscript{†}skylynfluke@gmail.com
\\
\textsuperscript{*}Equal contribution 
\textsuperscript{†}Corresponding author
}
}

\maketitle


\begin{abstract}
Existing EEG recognition models suffer from poor cross-paradigm generalization due to dataset-specific constraints and individual variability. To overcome these limitations, we propose \textsc{Bite} (Bidirectional Time-Freq Pyramid Network), an end-to-end unified architecture featuring robust multistream synergy, pyramid time-frequency attention (PTFA), and bidirectional adaptive convolutions. The framework uniquely integrates:
\textit{1) Aligned time-frequency streams maintaining temporal synchronization with STFT for bidirectional modeling,
2) PTFA-based multi-scale feature enhancement amplifying critical neural patterns,
3) BiTCN with learnable fusion capturing forward/backward neural dynamics.}
Demonstrating enhanced robustness, \textsc{Bite} achieves state-of-the-art performance across four divergent paradigms (BCICIV-2A/2B, HGD, SD-SSVEP), excelling in both within-subject accuracy and cross-subject generalization. As a unified architecture, it combines robust performance across both MI and SSVEP tasks with exceptional computational efficiency. Our work validates that paradigm-aligned spectral-temporal processing is essential for reliable BCI systems. 
Just as its name suggests, \textsc{Bite} ``takes a bite out of EEG.'' We publicly release the model as \textsc{BiteEEG}, and the source code is available at \href{https://github.com/cindy-hong/BiteEEG}{https://github.com/cindy-hong/BiteEEG}.
\end{abstract}


\section{Introduction}

Electroencephalogram (EEG)-based brain-computer interfaces (BCIs) offer a non-invasive, portable, and high-temporal-resolution pathway for direct neural communication and control, holding significant promise for rehabilitation, assistive technologies, and human-computer interaction~\cite{wang2024freaml}.
Motor imagery (MI) and steady-state visual evoked potentials (SSVEP) represent two dominant EEG paradigms due to their distinct neural signatures and practical utility. MI elicits characteristic modulations in sensorimotor rhythms through imagined movement, while SSVEP generates robust oscillatory responses to flickering visual stimuli~\cite{luo2022dual, liu2024eeg}.

Despite decades of advancement, a critical barrier hinders the real-world adoption of EEG-BCIs, specifically, poor generalization across diverse paradigms, experimental sessions, and individual subjects. EEG signals are inherently noisy, non-stationary, and exhibit substantial inter-subject and intra-subject variability~\cite{guo2023adaptive, brandl2015bringing}. Crucially, models optimized for one specific paradigm, such as a particular MI task, or dataset frequently exhibit significant performance degradation when applied to others like SSVEP~\cite{mcgeady2019hybrid}. This cross-paradigm gap necessitates frequent recalibration or paradigm-specific retraining, undermining BCI practicality and robustness.

\begin{figure}
    \centering
    \includegraphics[width=\linewidth]{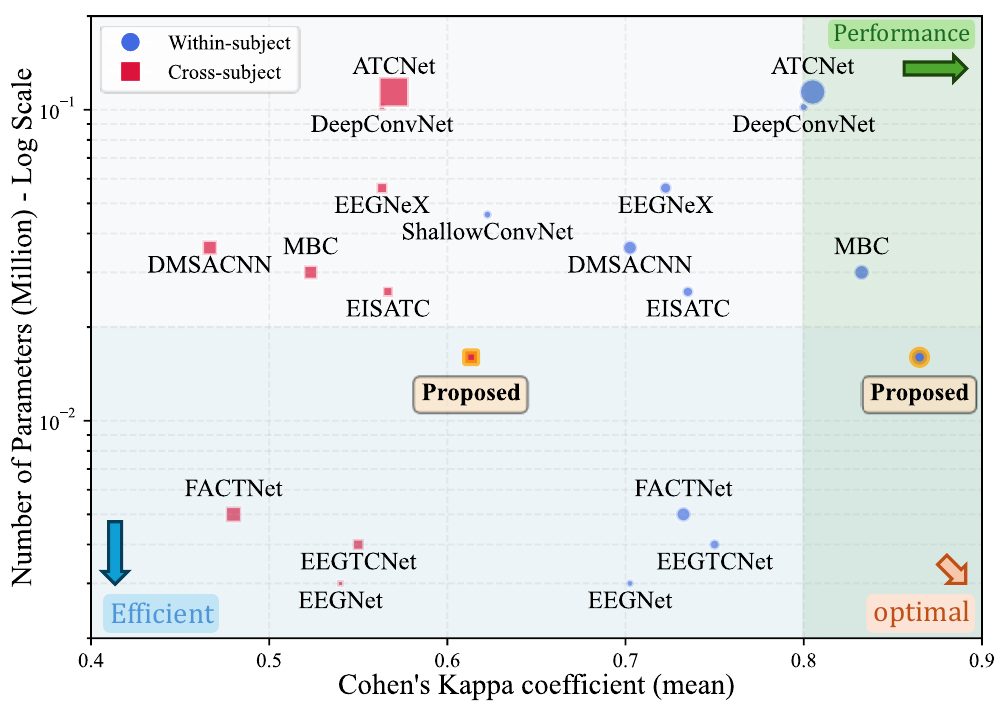}
    \caption{\textbf{Performance comparison across different approaches.} The bubble plot provides a visualization of the trade-offs among performance, efficiency, and training time (bubble size). The results emphasize evaluations across different paradigms. The reported results are a composite of four datasets to ensure robustness and generalizability.}
    \label{fig:abstract}
\end{figure}

Deep learning has revolutionized EEG analysis, automating feature extraction and achieving state-of-the-art results within constrained datasets~\cite{xiao2022efficient, liu2025recent}. However, existing architectures struggle with reliable cross-paradigm generalization. We identify three fundamental limitations.
\textit{1) Inadequate Representation of Time-Frequency Dynamics.} While dual-stream architectures have been proposed to process time and frequency domains, they often rely on Fast Fourier Transform (FFT) for frequency information. FFT provides a global frequency representation but fundamentally fails to capture the dynamic evolution of spectral components within localized time windows inherent in non-stationary EEG signals~\cite{ingale2014harmonic}. This limitation is critical for accurately modeling cortical activity during processes like MI.
\textit{2) Bidirectional Temporal Modeling Gap.} Neural processes underlying EEG patterns like MI involve complex temporal dependencies spanning preparation, execution, and recovery phases, with context potentially flowing both forwards and backwards~\cite{ingolfsson2020eeg}. Conventional recurrent like LSTM or convolutional like TCN architectures often model these dependencies with constrained receptive fields, missing crucial bidirectional context.
\textit{3) Limited Adaptive Multi-Scale Emphasis.} While attention mechanisms enhance relevant features~\cite{altaheri2022physics}, they often lack a principled way to leverage the hierarchical nature of time-frequency information. Existing methods rarely emphasize the most discriminative multi-scale spectral components to refine complementary temporal patterns across varying granularity.

To overcome these limitations and bridge the cross-paradigm gap, we introduce the \textbf{Bi}directional \textbf{T}ime-Frequency Pyramid \textbf{E}nhanced Network (\textsc{Bite}), a unified, robust, and computationally efficient architecture designed for cross-paradigm EEG classification. 
Our contributions are as follows.
\begin{itemize}

     \item \textbf{Unified Architecture for Diverse Paradigms.} \textsc{Bite} is an end-to-end framework handling fundamentally different EEG paradigms (e.g., MI, SSVEP) without paradigm-specific redesign or frequent recalibration, eliminating specialized models per BCI type.
    
     \item \textbf{Robust Time-Freq Synergy with Pyramid Attention.} \textsc{Bite} models the synergy between spectral dynamics using STFT and temporal signals. The Pyramid Time-Frequency Attention (PTFA) leverages multi-scale spectral features to guide attention on discriminative temporal patterns, ensuring robust cross-subject generalization.
    
     \item \textbf{Efficient Bidirectional Context Integration.} \textsc{Bite}'s Bidirectional Temporal Convolutional Network (BiTCN) captures long-range contextual dependencies through learnable adaptive fusion of forward/backward contexts, boosting efficacy for complex neural patterns with minimal computational overhead.
\end{itemize}

Through rigorous evaluation on four challenging and divergent EEG datasets, \textsc{Bite} demonstrates unprecedented cross-paradigm generalization, as shown in Fig.~\ref{fig:abstract}. 
\textsc{Bite} establishes that a unified architecture explicitly modeling dynamic time-frequency interactions and bidirectional context is essential for building reliable and generalizable BCI systems.

\section{Related Work}

\noindent\textbf{DL in BCI.} 

In recent years, deep learning (DL), particularly CNNs, has achieved promising results in EEG signal decoding research. Schirrmeister \textit{et al.}~\cite{schirrmeister2017deep} proposed two fundamental models. DeepConvNet employs a deep architecture with multiple convolution-pooling blocks and dense layers to process complex spatiotemporal features. This model demonstrated significant effectiveness in P300 experiments. In contrast, ShallowConvNet adopts a shallow design with only two convolutional layers. These layers perform temporal and spatial filtering respectively, mimicking the traditional FBCSP algorithm workflow. Comparative studies of these models indicate that architectural depth is not the determining factor in EEG signal processing. Rather, rational design principles prove more crucial. Subsequently, Lawhern \textit{et al.} introduced EEGNet~\cite{lawhern2018eegnet}. This design ingeniously integrates depthwise and separable convolutions, significantly improves efficiency and generalization across various BCI paradigms.

\noindent\textbf{TCN in BCI.} Temporal Convolutional Networks (TCNs) are architectures specifically designed for sequential data processing~\cite{bai2018empirical}. Building upon this foundation, Ingolfsson et al. (2020) introduced EEG-TCNet~\cite{ingolfsson2020eeg}. This model fuses TCN with EEGNet architecture. It utilizes dilated causal convolutions to construct exponentially growing receptive fields. The approach captures long-range temporal relationships while maintaining real-time performance, thereby addressing long-term temporal dependency issues in EEG signals.

\noindent\textbf{Attention in BCI.}
Attention mechanisms originated in natural language processing. Vaswani \textit{et al.}~\cite{vaswani2017attention} introduced these mechanisms through the Transformer architecture leading to rapid adoption. In EEG signal processing, attention mechanisms provide a powerful approach to highlight important features while suppressing noise information.
ATCNet~\cite{altaheri2022physics} pioneered the integration of multi-head self-attention mechanisms with TCN. By fusing attention mechanisms with sliding-window temporal convolutions, it enhanced the capacity to capture complex input data features. In 2024, EISATC-Fusion~\cite{liang2024eisatc} achieved significant performance improvements through a four-module synergistic strategy. This strategy comprises multi-scale temporal feature extraction, cosine similarity multi-head self-attention, parameter-efficient temporal modeling, and multi-level fusion mechanisms. In 2025, MBCNN-EATCFNet (MBCNet)~\cite{xiong2025mbcnn} captured spatiotemporal features at different scales by combining multi-branch and multi-scale structures. The model integrated Efficient Channel Attention (ECA) mechanisms into TCN, enabling bidirectional long-term temporal dependency extraction~\cite{wang2020eca}. These advances demonstrate the continuous optimization potential of attention mechanisms in the BCI field.

\noindent\textbf{Multistream in BCI.}
While the aforementioned deep learning models demonstrate effectiveness in extracting spatiotemporal features, they often overlook the importance of frequency-domain information. This oversight is particularly significant in neurophysiology, where frequency-domain information plays a crucial role in interpreting EEG signals. To address this issue, researchers began exploring methods to effectively integrate frequency-domain information into deep learning architectures.
In 2022, Huang \textit{et al.}~\cite{huang2022classification} proposed DCNN, which constructs a dual-stream convolutional neural network architecturen. This architecture processes time-domain and frequency-domain signals separately and combines time-frequency features through linear weighted fusion.
In 2024, Ke \textit{et al.}~\cite{ke2024fact} proposed FACTNet, which effectively integrates frequency-domain information through frequency-domain adaptive convolutional transformation. The model further introduces Frequency-domain Adapter (FA) modules and Time-domain Periodic Inception (TPI) modules. 

However, the above methods obtain frequency-domain information through FFT, which fundamentally maps entire time series to frequency-domain space and cannot capture frequency-domain dynamic changes within local time windows in EEG signals. This limitation becomes apparent when processing highly dynamic and non-stationary brain signals. In contrast, Short-Time Fourier Transform (STFT) maintains temporal resolution while preserving frequency-domain analysis capabilities through a sliding window mechanism. It better reflects the time-frequency dynamic characteristics of cortical activity during motor imagery processes.

Although the above approaches have their own limitations, these research advances show that multistream architectures combining TCN, and attention mechanisms have great potential to fully utilize frequency-domain information. 

\section{Methods}

\begin{figure*}[t!]
\centering
\includegraphics[width=0.86\textwidth]{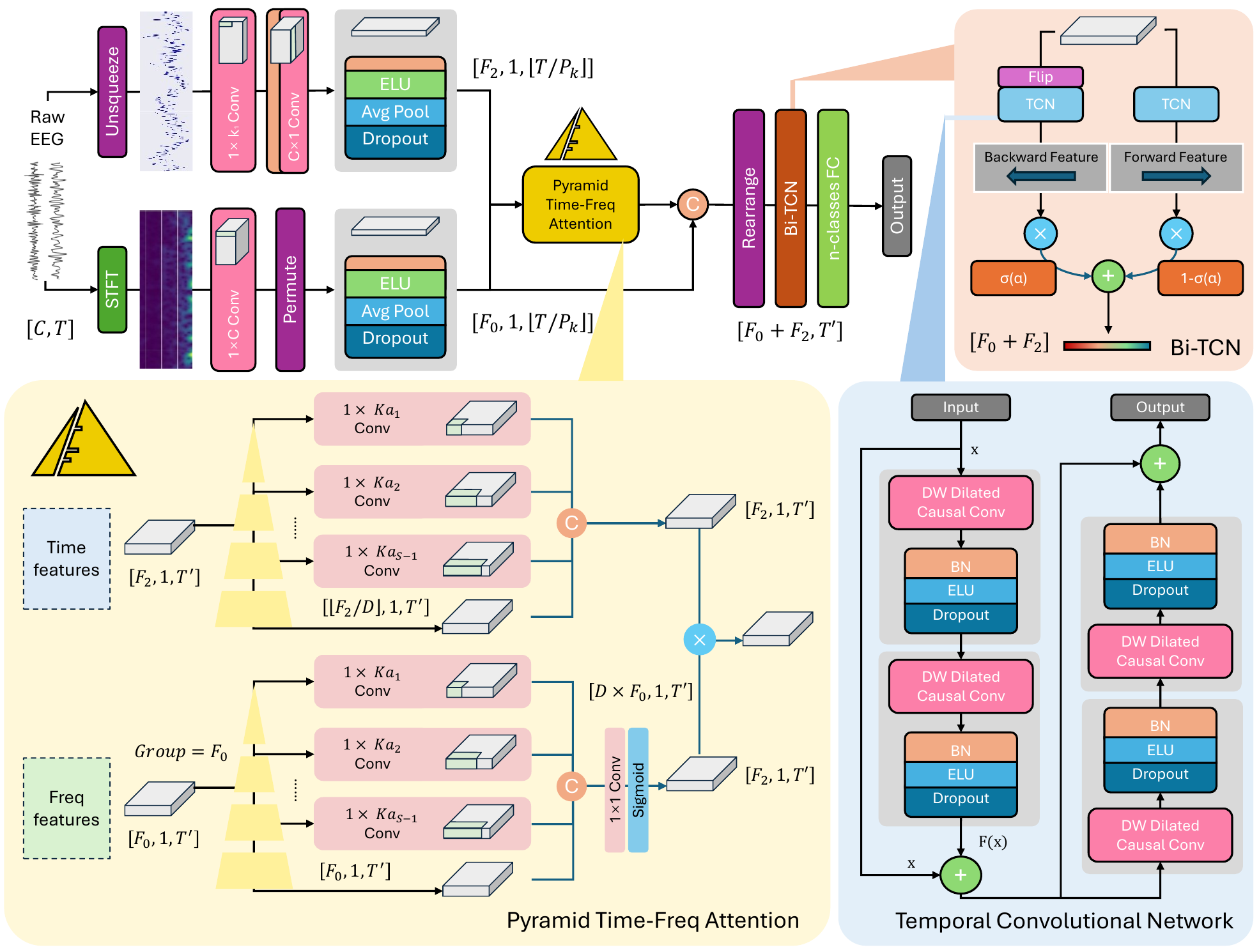}
\caption{Overview of the proposed \textsc{Bite} architecture including PTFA and Bi-TCN.}
\label{fig:architecture}
\end{figure*}

This section provides a comprehensive description of the proposed end-to-end framework, \textsc{Bite}, designed specifically for EEG signal classification. \textsc{Bite} leverages complementary information from both temporal and frequency domains of EEG signals through a dual-stream parallel architecture. The model incorporates multi-scale feature extraction, enhanced by pyramid attention mechanisms and bidirectional temporal modeling, to achieve robust brain pattern recognition. The core components include: (1) a dual-stream feature extraction network for separate processing of raw temporal and transformed frequency-domain inputs; (2) a pyramid time-frequency attention module (PTFA) for adaptive feature enhancement; (3) a bidirectional temporal convolutional network (BiTCN) for deep temporal sequence modeling; and (4) an adaptive feature fusion and classification mechanism. The subsequent subsections and Fig.~\ref{fig:architecture} elaborate on the detailed architecture and implementation of this framework.

\subsection{Dual-Stream Feature Extraction Network}
The first step is to extract both temporal and spatial features from raw EEG signals. The time domain is advantageous for capturing event-related potentials (ERPs), while the frequency domain allows analysis of oscillatory components. Combining both domains enables the network to extract features from multiple perspectives and prepares it for downstream classification in MI and SSVEP paradigms. Therefore, \textsc{Bite} adopts a dual-stream feature extraction structure.

Furthermore, to address cross-subject variability in EEG signals, we apply Euclidean Alignment (EA) \cite{he2019transfer} as a preprocessing step. This aligns raw EEG trials into a common reference space, enhancing model generalization across subjects. (detailed derivation in supplementary material) 

\subsubsection{Temporal Stream}
The temporal stream processes raw multi-channel EEG signals using a two-stage convolutional architecture to extract spatiotemporal features. The first stage employs one-dimensional temporal convolution to capture key temporal dynamics, while the second stage uses spatial convolution to learn cross-channel spatial patterns. The pipeline incorporates feature normalization, nonlinear activation, downsampling, and regularization to fully leverage the spatiotemporal structure of EEG signals. Mathematically, given an input EEG signal $X \in \mathbb{R}^{B \times 1 \times C \times T}$, where $B$ denotes batch, $C$ the electrode channel, $T$ the initial time samples, the two-stage temporal convolution architecture can be expressed as
\begin{align}
X^{(1)} &= \text{Conv}_{\text{time}}(X), \\
X^{(2)} &= \text{Conv}_{\text{space}}(X^{(1)}).
\end{align}
After normalization and activation, average pooling with factor \( P_k \) downsamples the time dimension to \( \lfloor T / P_k \rfloor \). Following dropout regularization, the temporal feature representation $F_{\text{time}} \in \mathbb{R}^{B \times F_2 \times 1 \times T'}$ is obtained, where $F_1$ is the number of filters in the temporal convolution, with $F_2 = F_1 \times D$ (depth multiplier $D$). This effectively isolates and models temporal and spatial characteristics, providing a robust foundation for downstream multimodal feature fusion and temporal modeling.

\subsubsection{Frequency Stream}
Frequency domain processing begins with Short-Time Fourier Transform, using Hann window with hop length 1 to maintain temporal alignment. Window lengths of 64 samples (MI) and 32 samples (SSVEP) match their respective temporal kernels, providing frequency resolutions of approximately 4Hz and 8Hz. The extracted frequency ranges (4-40Hz for MI, 8-64Hz for SSVEP) capture task-relevant neural oscillations: $\mu$ and $\beta$ rhythms for MI, and fundamental frequencies with harmonics for SSVEP.

The number of frequency bins $F_0$ within each extracted range is determined by the window length and sampling rate parameters, which form the frequency dimension of the reformulated tensor $X_{\text{freq}} \in \mathbb{R}^{B \times F_0 \times 1 \times T}$. This tensor is subsequently processed through grouped spatial convolution, where each frequency bin forms an independent group, enabling the network to learn frequency-specific spatial filters while maintaining parameter efficiency. Following normalization, activation, pooling, and dropout operations, the temporal feature representation $F_{\text{freq}} \in \mathbb{R}^{B \times F_0 \times 1 \times T'}$ is obtained.

\subsection{Pyramid Time-Frequency Attention Mechanism (PTFA)}
The PTFA module is a key component in \textsc{Bite} that bridges dual-stream features. Its core principle leverages frequency selectivity from frequency domain information to guide adaptive enhancement of temporal features. Crucially, it employs a symmetric multi-scale pyramid structure to capture long-range dependencies and multi-resolution correlations across the time-frequency domains.

\subsubsection{Multi-Scale Pyramid Structure}
PTFA adopts a symmetric multi-scale processing strategy, performing pyramid-based feature extraction at identical scales for both temporal and frequency domain features.  

The input temporal feature tensor $F_{\text{time}} \in \mathbb{R}^{B \times F_2 \times 1 \times T'}$ is partitioned into $D$ distinct channel groups. Each group contains $F_2/D$ channels. These groups are processed differently to capture multi-scale information: The initial group preserves the original temporal characteristics without modification; remaining groups $i$ (where $i = 2, 3, ..., D$) utilize convolution with kernel dimensions $(1, 2i-3)$. This design enables the model to simultaneously capture original features and multi-scale features.

The input frequency feature tensor $F_{\text{freq}} \in \mathbb{R}^{B \times F_0 \times 1 \times T'}$ (where $F_0$ corresponds to STFT frequency bands) is processed through $D$ parallel branches, each mirroring the scale of its temporal counterpart: The first branch maintains original frequency features; subsequent $D-1$ branches use grouped convolution (groups = $F_0$), where subsequent branches $i$ (for $i = 2, 3, ..., D$) use grouped convolution (groups = $F_0$) with kernel size $(1, 2i-3)$. Grouped convolution ensures independent processing of each frequency channel, while maintaining independence of frequency dimension.

\subsubsection{Cross-Domain Attention Generation}
PTFA generates attention weights for temporal features using the multi-scale spectral representations. The outputs from all $D$ frequency branches are concatenated along the channel dimension, producing an enriched spectral feature map.

\begin{equation}
\mathbf{F} = \text{Concat} \left( \mathbf{F}_{\text{freq}}, \left\{ \text{GroupConv}_{1\times(2i-3)}(\mathbf{F}_{\text{freq}}) \right\}_{i=2}^{D} \right),
\end{equation}
\begin{equation}
\mathbf{A} = \sigma \left(\text{Conv}_{1\times1}(\mathbf{F})\right),
\end{equation}
where $\sigma(\cdot)$ denotes the sigmoid activation. The attention map $\mathbf{A} \in \mathbb{R}^{B \times F_2 \times 1 \times T'}$ is guided by frequency features.

\subsubsection{Feature Enhancement Mechanism}
The enhanced temporal feature representation $\mathbf{Y}$ is generated by element-wise multiplication of the multi-scale temporal features $\mathbf{X}_{ms}$ with the frequency-derived attention map $\mathbf{A}$:
\begin{equation}
\mathbf{Y} = \mathbf{X}_{ms} \odot \mathbf{A} \in \mathbb{R}^{B \times F_2 \times 1 \times T'},
\end{equation}
where $\odot$ denotes the Hadamard product. 

This soft attention mechanism adaptively modulates temporal feature response strength through frequency information, achieving cross-modal feature enhancement. The output $\mathbf{Y}$ maintains the same dimensions as the input temporal feature $\mathbf{F}_{\text{time}}$, ensuring seamless integration with subsequent network layers and architectures.

\subsection{Bidirectional Temporal Convolutional Network (BiTCN)}

The concatenated time-frequency features are fed into the BiTCN for deep temporal modeling. Unlike unidirectional TCNs, BiTCN comprehensively captures temporal dependencies through independent forward and backward branches, making it particularly suitable for handling bidirectional temporal patterns prevalent in brain electrical signals.

\subsubsection{TCN Basic Architecture}
Each direction of the BiTCN consists of stacked residual TCN blocks. Each block employs two dilated causal convolution layers to expand the receptive field, enabling the model to capture long-range temporal dependencies efficiently. The residual connections mitigate gradient vanishing issues, enhancing training stability.

The convolution layers maintain consistent channel dimensions and share kernel sizes but differ in dilation rates, which grow exponentially ($2^i$, where $i$ is the layer index). This ensures efficient receptive field expansion with minimal parameter growth, allowing the model to capture hierarchical temporal patterns.

\subsubsection{Causal Convolution and Bidirectional Processing}
BiTCN employs bidirectional causal convolution to comprehensively model temporal dependencies in EEG signals. The forward branch employs causal convolution to ensure that the output at each time step depends solely on current and past inputs. This causality constraint aligns with the natural temporal sequence, particularly suitable for modeling anticipation and preparation processes in EEG signals. The backward branch achieves reverse causal modeling through temporal reversal. After flipping input features along the temporal dimension, they undergo identical causal convolution processing, enabling access to future information. This design is crucial for capturing after-effects and feedback mechanisms in EEG. Both branches use independent parameters to learn asymmetric temporal dependencies, enhancing expressiveness for EEG tasks.

\subsubsection{Temporal Feature Extraction Strategy}
BiTCN adopts a ``last-moment feature" extraction strategy. After multi-layer TCN processing, both forward and backward branches retain features from the last time step as the representation of the entire sequence. For the forward branch, the last time step integrates information from the entire sequence. For the backward branch, due to temporal reversal, its last time step corresponds to the original sequence's first moment, embedding information propagated backward. This approach avoids information loss from pooling operations while preserving temporal localization.

\subsection{Adaptive Feature Fusion}
\textsc{Bite} introduces learnable fusion weights to achieve adaptive bidirectional feature fusion. Let $H_f$ denote the forward branch output and $H_b$ denote the backward branch output. The fusion process is defined as:
\begin{align}
H = \alpha \cdot H_f + (1-\alpha) \cdot H_b, \end{align}
where $\alpha$ is a learnable scalar constrained to $[0,1]$ through a sigmoid activation. This design allows the model to adaptively balance forward and backward information based on task-specific characteristics, ensuring robust feature integration while maintaining training stability.

\section{Experiments}

To comprehensively evaluate the performance advantages of \textsc{Bite}, we designed a series of experiments comparing it with current state-of-the-art EEG recognition methods in both within-subject and cross-subject settings. Additionally, we conducted extensive analyses, including hyperparameter tuning, ablation studies, and efficiency evaluations, to validate the robustness, generalization, and computational efficiency of the proposed model.

\subsection{Experimental Setup}
\subsubsection{Datasets And Data Preprocessing}

We utilized two types of signals in our study: EEG signals from MI and SSVEP tasks. These datasets provide distinct signal characteristics and experimental paradigms, enabling comprehensive analysis for BCI research.

The BCI Competition IV-2A dataset~\cite{lawhern2018eegnet}, created by Graz University of Technology in 2008, is widely used in BCI research. It contains EEG data from nine subjects performing four MI tasks: left-hand, right-hand, both feet, and tongue. EEG signals were recorded from 22 Ag/AgCl electrodes at a sampling rate of 250Hz. Each subject underwent two sessions (training and testing) conducted on different days, with 288 artifact-free trials per session. The analysis window was from 2-6 seconds after the MI cue, providing 4-second time series (1,000 sampling points). 

The BCI Competition IV-2B dataset~\cite{lawhern2018eegnet} contains EEG data from nine subjects performing two MI tasks (left-hand and right-hand). Signals were recorded from three bipolar electrodes (C3, Cz, C4) and three  electrooculograms (EOG) electrodes at 250Hz. Each subject completed five sessions: two without feedback (60 trials per task) and three with visual feedback (80 trials per task). The analysis window was from 3-7 seconds after the MI prompt, yielding 4-second time series (1,000 sampling points).

The High-Gamma Dataset (HGD) \cite{schirrmeister2017deep} comprises EEG recordings from 14 healthy subjects performing four MI tasks: left-hand movement, right-hand movement, foot movement, and rest. Data were originally acquired at 500Hz via 128 electrodes and subsequently downsampled to 250Hz. Each subject completed roughly 1000 trials.

The SD-SSVEP dataset~\cite{masaki2015functional} contains EEG data from 10 subjects performing visual target gazing tasks with 12 frequency-modulated stimuli. EEG signals were recorded from eight occipital electrodes using the BioSemi ActiveTwo system (Biosemi, Inc.). Each subject completed 15 experimental blocks, with 180 trials in total, and each trial lasted for 4s. The EEG signals were sampled at a sampling rate of 2048Hz, and then downsampled to 256Hz. Since the estimated visual delay was 0.135s, the analysis window comprised the time window of [0.135s, 1.135s] after the stimulus onset, providing 256 sampling points.

\subsubsection{Baselines}
We compare our proposed model with several foundational deep learning architectures for EEG decoding, including DeepConvNet~\cite{schirrmeister2017deep}, ShallowConvNet~\cite{schirrmeister2017deep}, and EEGNet~\cite{lawhern2018eegnet}. Additionally, we evaluate against state-of-the-art temporal-convolutional models such as EEG-TCNet~\cite{ingolfsson2020eeg} and ATCNet~\cite{altaheri2022physics}, which integrate temporal dynamics and attention mechanisms. To further assess performance, we include recent advanced frameworks: EEGNeX~\cite{chen2024toward}, MBCNNEATCFNet~\cite{xiong2025mbcnn}, DMSACNN~\cite{liu2025dmsacnn}, EISATC~\cite{liang2024eisatc}, and frequency-domain approaches including FACTNet~\cite{ke2024fact}. These baselines represent diverse methodological approaches spanning convolutional networks, temporal modeling, attention mechanisms, and time-frequency processing to comprehensively contextualize our model’s capabilities in cross-subject and cross-session BCI tasks.

\subsubsection{Training Details}
All experiments utilized an NVIDIA A100 GPU (40GB VRAM). The software environment was based on the PyTorch 2.7.1 deep learning framework, running in Python 3.10.16 with CUDA 12.4. Training strategies were configured differentially according to task characteristics (within-subject vs. cross-subject) and signal types (MI vs. SSVEP). In the cross-subject experiments, a Leave-One-Subject-Out (LOSO) setting is adopted. Models were trained for the complete number of epochs by default. The random seed was set to \textbf{2025} for reproduction.

\subsection{Result and Analysis}
We conducted a systematic evaluation of the proposed \textbf{\textsc{Bite}} model on datasets from MI and SSVEP tasks, comparing it against various SOTA methods. To investigate the model's capability to effectively extract neural signals within a single subject and its generalization ability across subjects, we performed both \textbf{Within-Subject} and \textbf{Cross-Subject} experiments on each dataset. Two key metrics were adopted for performance evaluation: Accuracy and Kappa. Accuracy provides an intuitive measure of the model's ability to correctly classify data, while Kappa adjusts for random guessing, offering a more robust evaluation of the agreement between the model's predictions and the true labels.

\subsubsection{Within-Subject Experiments}
The main findings of the within-subject experiments are summarized in Table \ref{tab:performance_comparison_within}. The Proposed model achieved the best or near-best performance across all datasets, demonstrating exceptional performance on complex tasks such as HGD (85.34$\pm$7.63) and BCICIV-2A (88.37$\pm$6.01). The robustness of the Proposed model is further supported by its superior Kappa scores, which indicate consistent and reliable predictions. In contrast, ShallowConvNet exhibited poor performance on SSVEP tasks due to its architectural misalignment with the unique characteristics of SSVEP signals. Overall, deep learning models demonstrated a clear advantage across the majority of tasks.

\input{withinresult}

The main findings of the within-subject experiments are summarized in Table \ref{tab:performance_comparison_within}. The Proposed model achieved the best or near-best performance across all datasets, demonstrating exceptional capability on complex tasks such as HGD (85.34$\pm$7.63) and BCICIV-2A (88.37$\pm$6.01). The robustness of \textsc{Bite} is further supported by its superior Kappa scores, which indicate consistent and reliable predictions.

\begin{figure}[!t]
\centering
\includegraphics[width=0.95\linewidth]{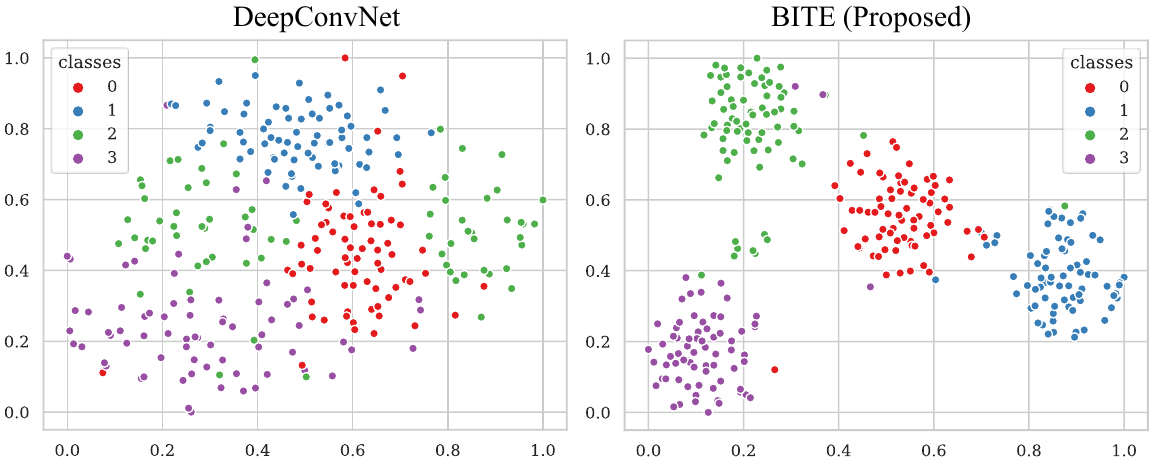}
\caption{\textbf{t-SNE of feature embeddings for Subject 3 in BCICIV-2A.} \textsc{Bite} exhibits superior separability between classes (distinct colors) and tighter clustering within classes compared to alternatives. \textit{Sub3 is selected due to convention.}}
\label{fig:tsne_bci}
\end{figure}

To qualitatively validate these results, we visualized the feature embeddings learned by the model via t-SNE dimensionality reduction. As illustrated in Fig. \ref{fig:tsne_bci} for Subject 3 in BCICIV-2A, the proposed method generates features with enhanced inter-class separation and intra-class compactness. This discriminative clustering elucidates the model's strong classification performance by demonstrating clear separation between distinct neural patterns while maintaining homogeneity within classes.

In contrast, ShallowConvNet exhibited significantly degraded performance on SSVEP tasks due to architectural incompatibility with the unique spectral characteristics of SSVEP signals. Overall, deep learning approaches demonstrated a clear advantage across the majority of tasks, with the proposed architecture achieving SOTA discriminability.

\subsubsection{Cross-Subject Experiments}
The main findings of the cross-subject experiments are presented in Table \ref{tab:performance_comparison_cross}. The Proposed model demonstrated strong generalization capabilities across all datasets, significantly outperforming other models on challenging tasks such as SD-SSVEP (79.72$\pm$19.43). However, the performance variability was higher due to the inherent complexity of the tasks and individual differences among subjects. Deep learning models (e.g., Proposed and EISATC) generally outperformed shallow models (e.g., ShallowConvNet), with shallow models often performing close to random guessing on difficult tasks.

\input{crossresult}

\input{within-sub-bite}

Analysis of the Decrement reveals insightful patterns about generalization capability. The proposed model exhibits superior performance decrements relative to within-subject experiments ($\Delta$ACC = 15.72, $\Delta\kappa$ = 0.22), indicating robust adaptation to distribution shifts between training and testing subjects. While shallow models like ShallowConvNet show smaller decrements ($\Delta$ACC = 8.03), this stems from limited representational capacity that yields near-chance performance on complex tasks rather than true generalization strength.

Further evaluation of model efficiency highlights the proposed architecture's practical advantages. With only 14 million parameters, it maintains an optimal balance between performance and computational requirements, significantly outperforming parameter-inefficient models like ATCNet (115M parameters) while delivering higher accuracy. This efficiency translates to substantially faster training times without compromising classification performance, as detailed in supplementary materials section.

\subsection{Hyperparameter Experiments}

\input{dropout}

To evaluate the impact of hyperparameter configurations on model performance, we conducted experiments by varying the dropout rate and kernel size of the TCN on the BCI Competition IV-2A dataset, focusing on within-subject classification tasks. 
Table \ref{tab:kernel_dropout} presents the average accuracy for different kernel sizes and dropout rates, ranging from 0.1 to 0.5. The results indicate that, across all kernel sizes, the performance generally peaked at a dropout rate of 0.3. A higher dropout rate led to excessive regularization, reducing accuracy, while lower rates resulted in insufficient regularization, also impairing performance. Furthermore, the kernel size significantly affected accuracy. As shown in Table \ref{tab:kernel_dropout}, a kernel size of 6 consistently achieved superior performance compared to other configurations. Smaller kernel sizes failed to capture adequate contextual information, while larger kernel sizes exhibited reduced accuracy, likely due to over-smoothing effects.

\subsection{Ablation Study}

\begin{figure}[ht]
\centering
\includegraphics[width=0.4\textwidth]{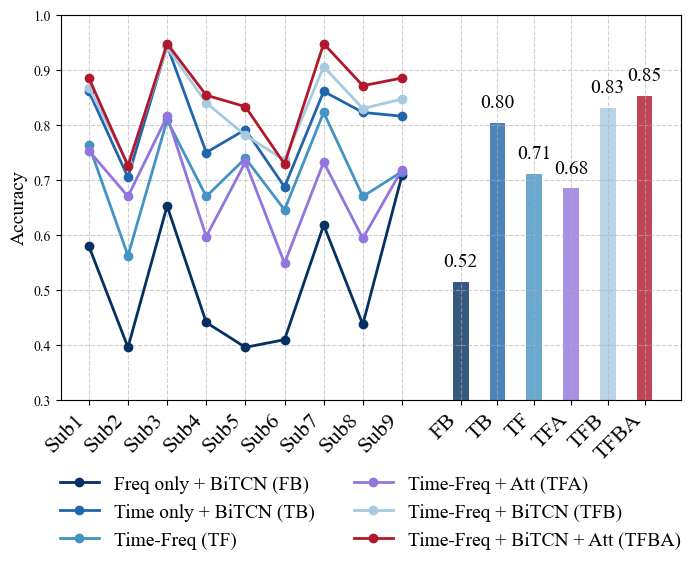}
\caption{\textbf{Ablation study results.} Ablation study results showing per-subject accuracy (solid lines) and mean accuracy (right-side bar chart) across configurations. Different configurations are represented by lines of different colors, and their mean accuracies are displayed as bars with matching colors.}
\label{fig:ablation}
\end{figure}

To evaluate the contribution of core components in our architecture, we conducted an ablation study with six configurations on BCI Competition IV-2A dataset, focusing on within-subject classification tasks. Performance was assessed using classification accuracy, with results shown in Fig.~\ref{fig:ablation}.

Using temporal features-only with BiTCN (TB) achieved a mean accuracy of 0.80, demonstrating moderate performance with low variance. In contrast, frequency features-only with BiTCN (FB) performed the worst, indicating limited robustness. Combining temporal and frequency features (TF) without BiTCN improved the mean accuracy to 0.71, highlighting their complementary nature. Adding Attention (TFA) slightly reduced accuracy to 0.68. Incorporating BiTCN (TFB) further enhanced performance. Finally, combining Time-Frequency features with both BiTCN and Attention (TFBA) achieved the highest accuracy of 0.85, demonstrating the synergistic benefits of BiTCN and attention for robust modeling.

In summary, the results clearly demonstrate that each module in the framework is necessary and effective. The collaboration of these modules significantly enhances performance.

\section{Conclusion}
This work addresses the critical challenge of cross-paradigm generalization in EEG-BCIs, where conventional models fail to maintain performance across diverse paradigms like motor imagery and steady-state visual evoked potentials due to EEG non-stationarity and inter-subject variability. We propose \textsc{Bite}, a unified architecture that bridges this gap through dynamic time-frequency representation and bidirectional context modeling. Experiment results validate \textsc{Bite} as a versatile, efficient solution for practical BCI deployment.
Future work will explore extending \textsc{Bite} to real-time, low-latency BCI systems and validating generalization across more paradigms.
By unifying robust feature learning and contextual modeling, \textsc{Bite} lays a foundation for reliable, plug-and-play BCIs.
\bibliographystyle{ieeetr}
\bibliography{BITE}

\end{document}

%% file: withinresult.tex
\begin{table*}[htbp]
\centering
\newcommand{\highlightfirst}[1]{\cellcolor[HTML]{FFD966}#1}
\newcommand{\highlightsecond}[1]{\cellcolor[HTML]{A9D0F5}#1}
\begin{tabular}{@{}c *{4}{ll} @{}}
\toprule
\textbf{Model}    & \multicolumn{2}{c}{\textbf{BCICIV-2A}} & \multicolumn{2}{c}{\textbf{BCICIV-2B}} & \multicolumn{2}{c}{\textbf{HGD}} & \multicolumn{2}{c}{\textbf{SD-SSVEP}} \\
\cmidrule(lr){2-3} \cmidrule(lr){4-5} \cmidrule(lr){6-7} \cmidrule(lr){8-9}
                  & \multicolumn{1}{c}{\textbf{Accuracy}} & \multicolumn{1}{c}{$\kappa$} & \multicolumn{1}{c}{\textbf{Accuracy}} & \multicolumn{1}{c}{$\kappa$} & \multicolumn{1}{c}{\textbf{Accuracy}} & \multicolumn{1}{c}{$\kappa$} & \multicolumn{1}{c}{\textbf{Accuracy}} & \multicolumn{1}{c}{$\kappa$} \\ \midrule
DeepConvNet       & 72.38±14.06** & 0.63±0.19      & \highlightsecond{85.51±8.85*}   & \highlightsecond{0.71±0.18}      & 93.08±4.57*  & 0.91±0.06      & \highlightfirst{95.50±7.15}    & \highlightfirst{0.95±0.08}      \\
ShallowConvNet    & 74.11±10.50** & 0.65±0.14      & 83.66±10.12** & 0.67±0.20      & 95.33±3.45   & \highlightsecond{0.94±0.05}      & 29.17±15.13*** & 0.23±0.17      \\
EEGNet            & 71.14±11.95** & 0.62±0.16      & 82.70±10.29*  & 0.65±0.21      & 93.97±2.73** & 0.92±0.04      & 64.83±20.93*** & 0.62±0.23      \\
EEGTCNet          & 78.43±9.56**  & 0.71±0.13      & 82.56±8.97**  & 0.65±0.18      & 94.67±3.51** & 0.93±0.05      & 73.33±24.46**  & 0.71±0.27      \\
ATCNet            & 81.02±7.75**  & 0.75±0.10      & 84.25±8.29**  & 0.69±0.17      & \highlightsecond{95.89±3.12}   & \highlightfirst{0.95±0.04}      & 84.50±19.51**  & 0.83±0.21      \\
EEGNeX            & 75.62±8.73**  & 0.67±0.12      & 82.87±7.21**  & 0.66±0.14      & 93.55±3.49*** & 0.91±0.05      & 67.83±24.92*** & 0.65±0.27      \\
MBCNet            & \highlightsecond{82.10±7.47}    & \highlightsecond{0.76±0.10}      & 84.79±7.14**  & 0.70±0.14      & 93.70±3.54*** & 0.92±0.05      & \highlightsecond{95.33±5.95}     & \highlightsecond{0.95±0.06}      \\
DMSACNN           & 77.59±11.16** & 0.70±0.15      & 84.65±8.95**  & 0.69±0.18      & 95.53±4.50   & \highlightsecond{0.94±0.06}      & 52.00±22.98*** & 0.48±0.25      \\
EISATC            & 78.13±9.38**  & 0.71±0.13      & 82.29±8.47**  & 0.65±0.17      & 92.95±4.05*** & 0.91±0.05      & 69.83±24.68*** & 0.67±0.27      \\
FACTNet           & 75.23±10.23** & 0.67±0.14      & 80.60±11.74*  & 0.61±0.23      & 93.63±3.44*** & 0.91±0.05      & 76.50±21.35*** & 0.74±0.23      \\
\textbf{Proposed} & \highlightfirst{85.34±7.63}    & \highlightfirst{0.80±0.10}      & \highlightfirst{88.37±6.01}    & \highlightfirst{0.77±0.12}      & \highlightfirst{95.93±3.16}   & \highlightfirst{0.95±0.04}      & 94.16±10.76    & 0.94±0.12      \\ \bottomrule
\end{tabular}
\caption{\textbf{Comparison of model performance in within-subject experiments across datasets.} Results are reported in terms of Accuracy (\%) and Kappa ($\kappa$). \textcolor{orange}{Orange cells} indicate the best performance, while \textcolor{blue}{blue cells} indicate the second-best performance for each metric. Asterisks denote the statistical significance of the differences between the proposed model and the baseline models: $^{*}p<0.05$, $^{**}p<0.01$, and $^{***}p<0.001$.}
\label{tab:performance_comparison_within}
\end{table*}    

%% file: crossresult.tex
\begin{table*}[htbp]
\centering
\newcommand{\highlightfirst}[1]{\cellcolor[HTML]{FFD966}#1}
\newcommand{\highlightsecond}[1]{\cellcolor[HTML]{A9D0F5}#1}

\begin{tabular}{@{}c *{3}{ll} ccc @{}}
\toprule
\textbf{Model}    & \multicolumn{2}{c}{\textbf{BCICIV-2A}} & \multicolumn{2}{c}{\textbf{BCICIV-2B}} & \multicolumn{2}{c}{\textbf{SD-SSVEP}} & \multicolumn{2}{c}{\textbf{Decrement}} & \multicolumn{1}{c}{\textbf{Parameters}} \\
\cmidrule(lr){2-3} \cmidrule(lr){4-5} \cmidrule(lr){6-7} \cmidrule(lr){8-9}
                  & \multicolumn{1}{c}{\textbf{Accuracy}} & \multicolumn{1}{c}{$\kappa$} & \multicolumn{1}{c}{\textbf{Accuracy}} & \multicolumn{1}{c}{$\kappa$} & \multicolumn{1}{c}{\textbf{Accuracy}} & \multicolumn{1}{c}{$\kappa$} & \textbf{$\Delta$ ACC} & \textbf{$\Delta$ $\kappa$} & (K) \\ \midrule
DeepConvNet       & 58.18±11.93** & 0.44±0.16      & 75.25±7.46       & 0.51±0.15      & \highlightsecond{76.50±22.05}       & \highlightsecond{0.74±0.24}      & 14.49 & 0.20      & 46 \\
ShallowConvNet    & 58.78±13.33** & 0.45±0.18      & 74.96±7.21       & 0.50±0.14      & 29.11±8.99***     & 0.23±0.10      & 8.03  & 0.12      & 20 \\
EEGNet            & 59.64±11.01*  & 0.46±0.15      & \highlightsecond{76.26±5.74}       & \highlightsecond{0.53±0.11}      & 65.89±24.07***    & 0.63±0.26      & 5.63  & 0.09      & 3 \\
EEGTCNet          & 60.86±7.81    & 0.48±0.10      & 76.14±6.37       & 0.52±0.13      & 68.00±22.65***    & 0.65±0.25      & 9.77  & 0.14      & 4 \\
ATCNet            & 61.30±10.88** & 0.48±0.15      & 75.95±5.65       & 0.52±0.11      & 73.11±23.09**     & 0.71±0.25      & 13.14 & 0.19      & 115 \\
EEGNeX            & 61.92±10.16*  & 0.49±0.14      & 74.18±6.14**     & 0.48±0.12      & 74.22±19.60**     & 0.72±0.21      & 5.33  & 0.10      & 55 \\
MBCNet            & 59.20±5.35*   & 0.46±0.07      & 74.18±7.62       & 0.48±0.15      & 65.94±19.74***    & 0.63±0.22      & 20.97 & 0.28      & 30 \\
DMSACNN           & 58.28±12.16** & 0.44±0.16      & 74.91±5.76       & 0.50±0.12      & 50.33±17.81***    & 0.46±0.19      & 10.24 & 0.16      & 21 \\
EISATC            & \highlightsecond{64.02±8.39}    & \highlightsecond{0.52±0.11}      & 75.80±6.65       & 0.52±0.13      & 69.06±19.65***    & 0.66±0.21      & 7.12  & 0.11      & 26 \\
FACTNet           & 56.92±11.45** & 0.43±0.15      & 76.01±5.86       & 0.52±0.12      & 53.56±18.24***    & 0.49±0.20      & 15.28 & 0.19      & 1 \\
\textbf{Proposed} & \highlightfirst{64.56±10.92}   & \highlightfirst{0.53±0.15}      & \highlightfirst{76.44±5.90}       & \highlightfirst{0.53±0.12}      & \highlightfirst{79.72±19.43}       & \highlightfirst{0.78±0.21}      & 15.72 & 0.22      & 14 \\ \bottomrule
\end{tabular}
\caption{\textbf{Comparison of model performance in cross-subject experiments across datasets.} Results are reported in terms of Accuracy (\%), Kappa ($\kappa$), their average decrements comparing to within-subject ($\Delta$), and parameter counts (K). \textcolor{orange}{Orange cells} indicate the best performance, while \textcolor{blue}{blue cells} indicate the second-best performance for each metric. Asterisks denote statistical significance: $^{*}p<0.05$, $^{**}p<0.01$, $^{***}p<0.001$. \textit{See full results in  supplementary material.}} 
\label{tab:performance_comparison_cross}
\end{table*}

%% file: within-sub-bite.tex
\begin{table*}[!ht]
\centering
\begin{tabular}{lcccc|cccccccccc}
\toprule
\textbf{Dataset} & \textbf{\#Sub} & \textbf{\#Channel} & \textbf{\#SP} & \textbf{\#Cat} & \textbf{Sub1} & \textbf{Sub2} & \textbf{Sub3} & \textbf{Sub4} & \textbf{Sub5} & \textbf{Sub6} & \textbf{Sub7} & \textbf{Sub8} & \textbf{Sub9} & \textbf{Sub10} \\
\midrule
BCICIV-2A & 9 & 22 & 1000 & 4 & 88.54 & 72.57 & 94.79 & 85.42 & 83.33 & 72.92 & 94.79 & 87.15 & 88.54 & — \\
BCICIV-2B & 9 & 3 & 1000 & 2 & 80.94 & 77.50 & 86.25 & 97.19 & 92.19 & 88.44 & 89.69 & 95.63 & 87.50 & — \\
HGD       & 14 & 44 & 1000 & 4 & 100.00 & 94.48 & 98.86 & 97.78 & 97.92 & 91.48 & 93.18 & 94.66 & 98.30 & 94.32 \\
SD-SSVEP    & 10 & 8 & 256 & 12 & 90.00 & 98.33 & 63.33 & 93.33 & 98.33 & 100.00 & 100.00 & 100.00 & 100.00 & 98.33 \\
\bottomrule
\end{tabular}
\caption{\textbf{Subject-wise within-Subject performance with dataset statistics.} \textit{See full results in supplementary material.}}
\label{tab:proposed_results}
\end{table*}

%% file: dropout.tex
\begin{table}[!ht]
\centering
\setlength{\tabcolsep}{7pt}  
\renewcommand{\arraystretch}{0.85}  
\footnotesize  
\newcommand{\highlightor}[1]{\cellcolor[HTML]{FFD966}#1}
\begin{tabular}{@{}c|cccccc@{}}
\toprule
\textbf{Kernel} & \multicolumn{6}{c}{\textbf{Dropout Rate}} \\
\cmidrule(lr){2-7}
 & 0.1 & 0.2 & 0.3 & 0.4 & 0.5 & Avg \\ 
\midrule
3 & \highlightor{0.8264} & 0.8210 & 0.8171 & 0.8098 & 0.7963 & 0.8141 \\ 
6 & 0.8395 & 0.8407 & \highlightor{0.8534} & 0.8291 & 0.8276 & 0.8380 \\ 
9 & 0.8295 & 0.8291 & \highlightor{0.8380} & 0.8318 & 0.8241 & 0.8291 \\ 
12 & 0.8225 & 0.8384 & 0.8395 & \highlightor{0.8407} & 0.8248 & 0.8346 \\ 
\bottomrule
\end{tabular}
\caption{\textbf{Accuracy by kernel size for different dropout rates.} Best performance is highlighted in \textcolor{orange}{orange}.}
\label{tab:kernel_dropout}
\vspace{-1em}
\end{table}